\documentclass[sigconf, 10pt]{acmart}
\usepackage{hhline}
\usepackage{booktabs} 
\usepackage{amsmath,amsfonts}
\usepackage{algorithmic}
\usepackage{graphicx}
\usepackage{textcomp}
\usepackage{xcolor}
\usepackage{float}
\usepackage{stfloats} 
\usepackage[T1]{fontenc}
\usepackage{scalerel}
\usepackage{amsmath}
\usepackage{verbatimbox}
\usepackage{multirow}
\usepackage{fdsymbol}
\usepackage{threeparttable}

\newcommand*{\mybulletshape}{\raisebox{-0.3ex}{\scalebox{1.8}{$\bullet$}}}
\DeclareRobustCommand{\mybluebullet}{\textcolor[HTML]{0066FF}{\mybulletshape}}
\DeclareRobustCommand{\myyellowbullet}{\textcolor[HTML]{FFC009}{\mybulletshape}}
\DeclareRobustCommand{\mygreenbullet}{\textcolor[HTML]{15B01A}{\mybulletshape}}

\acmYear{2025}\copyrightyear{2025}
\setcopyright{cc} \setcctype{by-nc}
\acmConference[SEC '25]{The Tenth ACM/IEEE Symposium on Edge Computing}{December 3--6, 2025}{Arlington, VA, USA}
\acmBooktitle{The Tenth ACM/IEEE Symposium on Edge Computing (SEC '25), December 3--6, 2025, Arlington, VA, USA}
\acmDOI{10.1145/3769102.3772713}
\acmISBN{979-8-4007-2238-7/25/12}

\acmSubmissionID{sec25-p15}

\begin{document}

\title{Efficient SAR Vessel Detection for FPGA-Based On-Satellite Sensing}
\author{Colin Laganier}
\authornote{Corresponding author: claganier@turing.ac.uk}
\orcid{0009-0005-7421-8875}
\affiliation{%
    \institution{The Alan Turing Institute}
    \city{London}
    \country{United Kingdom}
}
\author{Liam Fletcher}
\orcid{0000-0003-4840-2474}
\affiliation{%
    \institution{The Alan Turing Institute}
    \city{London}
    \country{United Kingdom}
}
\author{Elim Kwan}
\orcid{0000-0001-6607-086X}
\affiliation{%
    \institution{The Alan Turing Institute}
    \city{London}
    \country{United Kingdom}
}
\author{Richard Walters}
\orcid{0000-0002-1704-8727}
\affiliation{%
    \institution{The Alan Turing Institute}
    \city{London}
    \country{United Kingdom}
}
\author{Victoria Nockles}
\orcid{0009-0002-7159-5980}
\affiliation{%
    \institution{The Alan Turing Institute}
    \city{London}
    \country{United Kingdom}
}

\renewcommand{\shortauthors}{Laganier et al.}

\begin{abstract}
Rapid analysis of satellite imagery within minutes-to-hours of acquisition is increasingly vital for many remote sensing applications, and is an essential component for developing next-generation autonomous and distributed satellite systems. On-satellite machine learning (ML) has the potential for such rapid analysis, by overcoming latency associated with intermittent satellite connectivity to ground stations or relay satellites, but state-of-the-art models are often too large or power-hungry for on-board deployment.
Vessel detection using Synthetic Aperture Radar (SAR) is a critical time-sensitive application in maritime security that exemplifies this challenge. SAR vessel detection has previously been demonstrated only by ML models that either are too large for satellite deployment, have not been developed for sufficiently low-power hardware, or have only been tested on small SAR datasets that do not sufficiently represent the difficulty of the real-world task.
Here we systematically explore a suite of architectural adaptations to develop a novel YOLOv8 architecture optimized for this task and FPGA-based processing. We deploy our model on a Kria KV260 MPSoC, and show it can analyze a $\sim$700 megapixel SAR image in less than a minute, within common satellite power constraints (\textless 10W). Our model has detection and classification performance only $\sim$2\% and 3\% lower than values from state-of-the-art GPU-based models on the largest and most diverse open SAR vessel dataset, xView3-SAR, despite being $\sim$50 and $\sim$2500 times more computationally efficient. This work represents a key contribution towards on-satellite ML for time-critical SAR analysis, and more autonomous, scalable satellites.\footnote{Code: \url{https://github.com/alan-turing-institute/sar-vessel-detection-fpga}}
\end{abstract}



\received{20 June 2025}
\received[revised]{8 October 2025}
\received[accepted]{19 October 2025}

\maketitle

\section{Introduction}
Rapid analysis of low earth orbit satellite imagery within hours of data acquisition is vital for a diverse range of remote sensing applications, from wildfire detection, to agricultural monitoring, to disaster response \cite{Ban2020,Sabir2024,Boccardo2015}, and latency requirements for many applications are moving beyond near real-time (1-3 hours), to quasi real-time (\textless 30 mins), with target latencies of \textless 5 minutes in some cases \cite{Kerr2020}. Furthermore, these shortest latency requirements will be increasingly important due to growing interest in autonomous and distributed satellite capabilities \cite{Araguz2018}, for example ``tip-and-cue'' processes \cite{Cudzilo2012} (where one satellite uses initial analysis to request further imagery from another satellite). Such autonomous tasking and inter-satellite coordination capabilities have been identified as critical for next-generation Earth observation systems \cite{Thangavel2024}, with small test systems already in orbit (e.g. NASA's Starling satellite swarm \cite{Gardill2023}).
But achieving such rapid analysis is challenging, due to 1) ever-increasing dataset sizes of large-area, high-resolution imagery (several gigabytes for individual images, missions often producing TB/day \cite{OECD2019}), 2) bottlenecks in both transfer of data to ground stations and then on to cloud-based compute (typically hours for each \cite{Tao2023}), and 3) the computational complexity of the analysis tasks themselves, which often require accurate extraction, detection or classification of small-magnitude signals in large, high-dimensional, diverse and noisy datasets (e.g. \cite{Ma2023B}).
On-satellite Machine learning (ML) has major potential to reduce the time from data acquisition to analysis, avoiding the latency associated with downlinking full satellite images (e.g. \cite{Gardill2023}). This latency includes delays of up to several hours for a satellite to pass over a given ground station \cite{Vasisht2021}, which cannot be reduced by ground station-based compute, and is challenging to reduce beyond tens of minutes even with deployment of additional multi-million USD ground stations \cite{Vasisht2021} or geostationary data relay satellites (residual latencies of up to $\sim$45 minutes \cite{EDA_EDRS} due to visibility between the observing and geostationary satellites).
However, on-satellite ML requires careful design and optimization of ultra-efficient processing architectures that can achieve required performance and speed of analysis, whilst also operating within severely restricted power budgets (typically ranging from 2 to 20 watts in CubeSats \cite{Heidt2000}), with limited computational resources, and on a constrained selection of edge processing hardware for which space-hardening is possible (e.g. FPGAs, ASICs).
Therefore, demonstration of ML architectures suitable for on-satellite deployment has to-date only been achieved for a handful of remote sensing applications (e.g. \cite{Maskey2020,Rapuano2021,Giuffrida2022,Ruzicka2023}). In this study we tackle this problem for vessel detection with Synthetic Aperture Radar (SAR) satellite data, a critical maritime security application for which on-satellite ML analysis is not yet possible but which exemplifies the deployment requirements and challenges. In particular, the ship detection task is computationally challenging, maritime intelligence is typically required within 30 minutes of collection to be considered actionable \cite{MicroSAR}, and tip-and-cue or other forms of reactive rapid tasking are highly relevant for detection and tracking of moving vessels.
SAR vessel detection has previously been demonstrated only by ML models that either are too large for satellite deployment \cite{Meng2024, Cao2024, Zhao2024, Yasir2024, Zhu2024}, have not been developed for sufficiently low-power hardware \cite{Gao2022}, or have only been developed and tested on small SAR datasets that do not sufficiently represent the complexity of the real-world task \cite{Yang2022, Huang2024, Fang2025}.
In this study, we make a major step towards real-world on-satellite deployment of ML models for SAR vessel detection. Our main contributions are:
\begin{itemize}
    \item We systematically explore a suite of previous architectural adaptations to develop a novel YOLOv8 architecture specifically optimized for SAR vessel detection and FPGA-based processing. 
    \item We deploy our model on a AMD/Xilinx Kria KV260 FPGA, and show it can analyze a 40,000 km$^2$ SAR scene ($\sim$700 megapixels) in less than a minute, within common satellite power constraints (\textless10W). 
    \item Our model is between $\sim$50 and $\sim$2500 times more computationally efficient than state-of-the-art GPU-based models, whilst achieving detection and classification performance only $\sim$2\% and 3\% lower than values from these models on the largest and most diverse open SAR vessel dataset, xView3-SAR \cite{Paolo2022}.
\end{itemize}

\section{Background and Related Work}

\subsection{Vessel Detection with Synthetic Aperture Radar and Machine Learning}\label{subsec:background}

The use of Synthetic Aperture Radar (SAR) imagery to detect ``dark'' ships (those that disable AIS; Automatic Identification Systems) is critical for maritime surveillance; including for monitoring and combating illegal fishing \cite{Kurekin2019}, piracy \cite{Renga2011}, and trafficking \cite{Tanveer2025}. These applications typically require detection and classification of ships in quasi real-time for rapid response, ideally within minutes of data acquisition \cite{Leong2025}.

\begin{figure*}[t]
    \includegraphics[width=\textwidth]{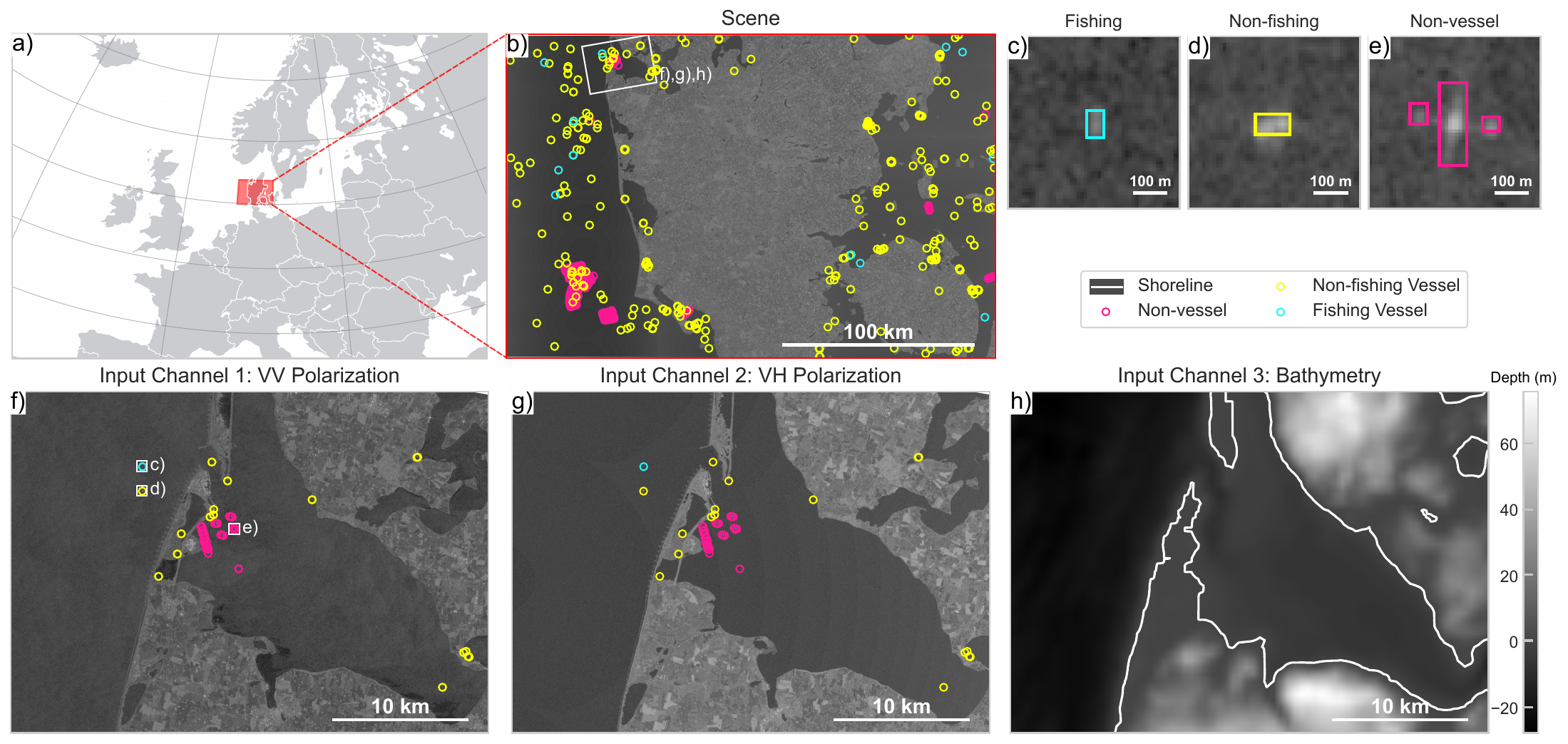}
    \caption{Example imagery from our dataset to illustrate the SAR Vessel Detection Task. (a,b) Single Sentinel-1 SAR scene over Denmark from the xView3-SAR dataset \cite{Paolo2022}. (b) shows amplitude of the VH band for the full scene with colored circles denoting ground truth fishing vessel (cyan), non-fishing vessel (yellow) and non-vessel (magenta) objects. (c-e) show sample close-up images of fishing vessel (c), non-fishing vessel (d) and non-vessel objects (e). (f-h) show the three channels used as model input in this study, for the highlighted region in (b); VV band SAR amplitude (f), VH band SAR amplitude (g) and bathymetry (h).}
    \label{fig:SAR_satellite}
\end{figure*}

SAR satellites deployed in near-polar low earth orbit are highly suited for this task due to their all-weather, 24-hour, active imaging capabilities, and their combination of large imaging footprint (single image up to several hundred kms wide, e.g. see Fig. \ref{fig:SAR_satellite}a-b) with moderate-high resolution (pixel size of meters to 10s of meters) \cite{Moreira2013}. These capabilities directly relate to how SAR satellites collect data; actively illuminating a large ground area with microwave pulses and recording the complex radar echoes (termed Level 0 or raw data). An initial processing step then leverages frequency modulation of the radar pulse and successive overlapping radar echoes to ``focus'' the image and increase the ground resolution. The resultant high-resolution images (Level 1 data) contain amplitude and phase data for one or multiple polarizations. Modern SAR satellites such as Sentinel-1, ALOS-4 and NISAR \cite{ESA2012,JAXA_ALOS4,NISAR2018} each generate multiple terabytes of imagery every day.

Vessel detection in SAR amplitude images is superficially similar to small object detection in RGB imagery, but aspects of the task and unique data modality make it significantly more challenging. In particular, vessels are much smaller than typical ``small objects''; average vessels in our dataset (Fig. \ref{fig:SAR_satellite}) are 9 pixels long and have area just 0.002\% relative to the size of an 800$\times$800 pixel chip, whilst the smallest vessels are 2 pixels long and have relative area 0.0002\%. These relative sizes are $\sim$40--400 times smaller than the lower bound of a commonly-used size definition for small objects in computer vision; 0.08--0.58\% of image area \cite{Chen2016}. Moreover, the vessels are similar in spatial extent and amplitude to background clutter (e.g. see Fig. \ref{fig:SAR_satellite}c-e), the inherent speckle noise in SAR images and high variability in vessel signatures across different sea states further complicates separation of vessels and background \cite{awais2025surveysarshipclassification}, and these challenges are especially pronounced in near-shore regions, making the task even harder in these areas \cite{Li2023}. Finally, the input has only two bands (dual-polarization) rather than three (RGB).

Computer vision techniques for SAR vessel detection \cite{Zhu2021} have addressed the challenges associated with task difficulty, enabling automated, adaptable and robust analysis and improving on accuracy and adaptability over traditional rule-based methods \cite{Yasir2023, Khalid2024}. 
But although state-of-the-art models achieve high accuracy across diverse maritime conditions (e.g. as represented by the largest and most diverse open SAR vessel dataset; xView3-SAR \cite{Paolo2022}), this has relied on large-scale architectures and ensemble methods. Such models are therefore too large and power-hungry for satellite deployment (the top 5 performing models from the competition associated with the xView3-SAR dataset have 89M to 754M parameters; 249 MB to 3.0 GB \cite{Paolo2022}; see blue symbols in Table \ref{tab:yolo_performance} and Fig. \ref{fig:xview3_models}). 
Various lightweight models for SAR vessel detection have also been proposed, using a range of architectures such as LSR-Det \cite{Meng2024}, SAR-Net \cite{Cao2024}, and YOLO \cite{Zhao2024, Yasir2024, Zhu2024}, but the majority of these models have either not been deployed or evaluated on edge hardware \cite{Meng2024, Cao2024, Zhao2024, Yasir2024, Zhu2024}, or else have been developed for and tested on platforms with power requirements exceeding satellite constraints \cite{Gao2022}.
Finally, the only models that have been deployed on hardware with low enough power to meet these constraints (Field-Programmable Gate Arrays; FPGAs \cite{Yang2022, Huang2024, Fang2025}) have only been trained and evaluated on relatively small and less representative SAR vessel datasets \cite{Zhang2021, Wang2019C, Wei2020} that do not reflect the real-world scale, diversity, and complexity of operational maritime scenarios captured in large-scale benchmarks such as xView3-SAR (1-2 orders of magnitude more ship instances, 2-3 orders of magnitude more image pixels).

\subsection{YOLOv8 for SAR \& Small Object Detection}\label{subsec:yolo_background}

In this study we use the lightweight YOLOv8 architecture \cite{Redmon2016} (Fig.~\ref{fig:yolov8_architecture}) as our base model, due to its previous success for both SAR vessel detection \cite{Luo2024, He2024} and small object detection \cite{Khalili2024} tasks. Here we briefly describe the base architecture and review existing literature on using YOLOv8 for SAR vessel and small object detection.
YOLOv8 adds anchor-free detection and has improved accuracy and processing speed relative to previous versions. It uses a modified CSPDarkNet53 architecture \cite{Bochkovskiy2020} as a feature extractor (Fig. \ref{fig:yolov8_architecture}a), which incorporates five downsampling operations from $2\times$ to $32\times$ and produces the scaled output feature maps P1-P5. This backbone uses a ``C2f'' block; an optimized Cross Stage Partial (CSP) block with two Conv modules (\textit{Convolution-BN-SiLU}) and \textit{n} bottlenecks. The SPPF (Spatial Pyramid Pooling - Fast) module then aggregates multi-scale spatial information in the final stage of the backbone, and the P3, P4, and P5 feature maps (of dimensions $100\times100$, $50\times50$, and $25\times25$, respectively, for a $800\times800$ pixel input) are passed to the model's neck. This neck (Fig. \ref{fig:yolov8_architecture}b) combines these multi-scale feature maps in both a top-down and bottom-up network structure, using a Path Aggregation Network Feature Pyramid Network (FPN) \cite{Lin2016,Liu2018}. Finally, the YOLOv8 heads (Fig. \ref{fig:yolov8_architecture}c) separate object classification (using binary cross-entropy loss) from bounding box regression (using a combination of Distribution Focal Loss (DFL) and Complete IoU (CIoU)).

\subsubsection{Modifications for SAR vessels and small objects:} Prior work has introduced four major groups of architectural and functional modifications to improve small target detection. First, modified convolutional blocks have been employed to enhance detection performance, to reduce computational complexity, or both. \cite{Luo2024} implemented both GhostConv and RepGhost bottlenecks to reduce model size without compromising representational capacity, whereas \cite{Bui2025,Xiao2023} replaced C2f blocks with C3 blocks, incorporating GhostConv bottlenecks to enhance the efficiency of small object detection. \cite{He2024} utilized Partial Convolution (PConv) as an alternative approach for improving feature learning efficiency on SAR scenes. 
Second, FPN modifications can also improve SAR detection. To preserve fine-grained spatial details essential for small target identification, \cite{Guo2024,Kang2024} incorporated the P2 feature map, which captures higher-resolution features lost in deeper layers. \cite{Guo2024} further added an Asymptotic Feature Pyramid Network (AFPN) to improve multi-scale feature fusion, while \cite{Khalili2024} implemented a Generalized Feature Pyramid Network (GFPN) for small-object detection in non-SAR contexts. 
Third, attention mechanisms have been adopted to improve feature discrimination in cluttered SAR imagery. \cite{Luo2024} implemented ShuffleAttention to model channel-wise dependencies, and both \cite{He2024, Guo2024} employed Efficient Multi-scale Attention (EMA \cite{Ouyang2023}) modules to enhance cross-scale feature representation. \cite{Feng2024} demonstrated improved localization of small objects in non-SAR images by applying Coordinate Attention (CA \cite{Hou2021}) to embed precise positional information within channel attention. 
Fourth, alternative loss functions have also been used to enhance localization accuracy for vessel detection, including Wise IoU (WIoU) loss to dynamically adjust focus based on anchor quality through a non-monotonic penalty mechanism \cite{Luo2024}, and MP-DIoU loss to improve overlap estimation by minimizing distances between the corners of predicted and ground truth boxes \cite{He2024}. The Powerful IoU (PIoU) loss has also been used to reduce localization error for non-SAR small-objects \cite{Khalili2024}, minimizing the Euclidean distance between bounding box corners.

\section{Methodology}
 In Section \ref{subsec:yolo_proposal} we describe our proposed model architecture, which employs a tailored set of modifications to adapt the YOLOv8 model to our specific task. We then detail our FPGA deployment in Section \ref{subsec:method_fpga}, followed by an overview of the xView3-SAR dataset in Section \ref{subsec:results_dataset}. Finally, we describe our evaluation metrics in Section \ref{subsec:results_metrics} and the training setup for our models in Section \ref{subsec:results_training}.

\begin{figure*}[t]
    \centering
    \includegraphics[width=.9\textwidth]{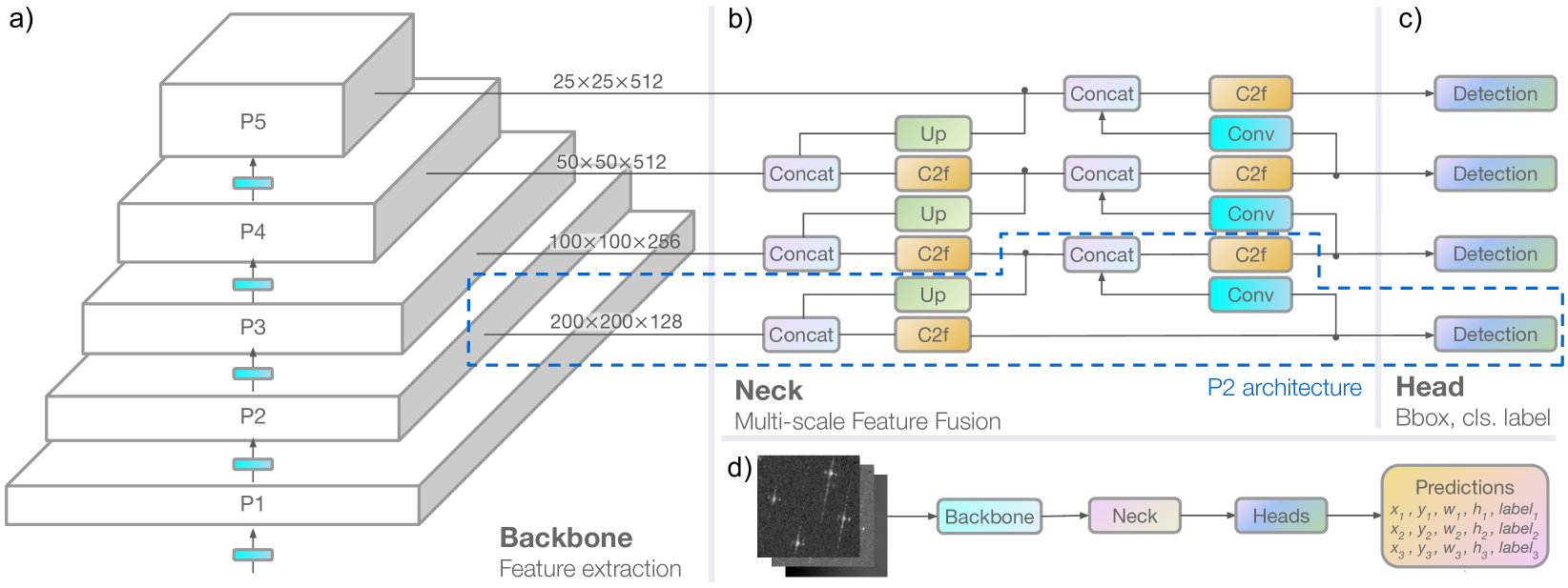}
    \caption{Architecture diagram of the YOLOv8-based object detection model. (a) feature extraction backbone, (b) multi-scale feature fusion neck, (c) detection head, (d) simplified end-to-end detection pipeline from SAR and bathymetry input to predictions. P2 architecture branch is shown in dashed blue box. Diagram modified from \cite{RangeKing2023}.}
    \label{fig:yolov8_architecture}
\end{figure*}

\subsection{Proposed Architecture}\label{subsec:yolo_proposal}

Our proposed architecture represents a novel combination of three of the modifications outlined in the previous section, to enhance SAR vessel detection performance and computational efficiency. In our experiments we systematically tested a broader set of the modifications discussed in the previous section; for brevity we present illustrative results showing why they were not incorporated in our final architecture in Table \ref{tab:yolo_performance} and briefly discuss these in Section \ref{subsec:results_yolo}.

First, to better leverage multi-scale feature representations, we extend the FPN by incorporating the P2 feature map and an additional detection head (Fig.~\ref{fig:yolov8_architecture}). Higher-resolution feature maps (P1, P2) better capture fine-grained details useful for detecting small targets, whilst deeper features (P4, P5) are more effective at detecting larger structures.

Second, we introduce two architectural improvements to enhance performance while reducing computational overhead. We replace the C2f blocks with CSP blocks featuring three Conv modules (referred to as C3), which incorporate an additional Conv module and enhance feature representation with minimal computational overhead. We also substitute the up- and down-sampling Conv blocks in the backbone and neck, as well as those in the bottleneck modules, with GhostConv \cite{Han2020}. This lightweight convolution variant reduces both computational and parameter overhead by generating feature maps through a two-stage process; a primary convolution produces a low-dimensional representation ($\lfloor c_{out}/2\rfloor$ channels), which is then concatenated with the output of a secondary depthwise convolution of matching dimensions.

Third, to address the inherent challenges of small object detection, we replace the standard IoU loss with PIoU2 \cite{Liu2024}. This modified loss function features an adaptive penalty factor based on target size (calculated from the average absolute distance between corresponding edges of prediction and ground truth boxes) and a non-monotonic focusing mechanism that concentrates learning on bounding boxes with moderate prediction quality. This approach has not previously been used for SAR vessel detection.

\subsection{FPGA Deployment}\label{subsec:method_fpga}\label{subsec:fpga_proposal}
In this study we specifically investigate FPGA deployment, due to their power efficiency and existing widespread use onboard satellites. We focus on a single FPGA device as the one selected fits within the strict power constraints of a representative small satellite platform.
We evaluate our proposed solution on the AMD/Xilinx Kria KV260 board, which integrates programmable logic (PL; Table \ref{tab:fpga_resources}) with a processing system (PS) based on a Quad-core Arm Cortex-A53 SoC. For model deployment, we utilized the Vitis AI 3.5 framework and created a custom hardware design targeting the PL of the Kria KV260 and incorporating the Deep Learning Processing Unit (DPU) 4.1 IP using Vivado 2023.2. The DPU is a programmable engine optimized for deep neural networks, which is leveraged by the Vitis AI instruction set to run models in the PL. In our design, the DPU is configured with one B4096 core and a clock frequency of 300~MHz. We quantized the model to INT8 using the Vitis AI Quantizer library, using symmetric quantization with a power-of-two scaling scheme (required configuration for the DPU), allowing for both Post-Training Quantization (PTQ) and Quantization-Aware Training (QAT) using a straight-through estimator \cite{Jain2020}. The DPU of the Kria KV260 platform (DPUCZDX8G) does not support the SiLU activation function used in YOLOv8, and so we modified the model to use Hard Swish activations \cite{Howard2019} instead. Finally, we deployed the compiled model on the FPGA using the Vitis AI Runtime, and evaluated model performance in terms of speed, power consumption, and accuracy.

\begin{table}[h]
    \centering
    \caption{FPGA resource utilization summary.}
    \label{tab:fpga_resources}
    \renewcommand{\arraystretch}{.95}
    \begin{tabular}{@{}ccccc@{}}
    \toprule
    \textbf{LUT} & \textbf{FF} & \textbf{DSP} & \textbf{BRAM} & \textbf{URAM} \\ 
    \midrule
    \multicolumn{5}{c}{\textit{Kria KV260}} \\
    117,129 & 234,240 & 1,248 & 144 & 64 \\ 
    \midrule
    \multicolumn{5}{c}{\textit{Vitis AI B4096 DPU}} \\
    62,104 & 123,486 & 710 & 109 & 40 \\
    (53.0\%) & (53.2\%) & (56.9\%) & (75.7\%) & (62.5\%) \\ 
    \midrule
    \multicolumn{5}{c}{\textit{Post-Processing Decoding Kernel}} \\
    45,671 & 32,011 & 188 & 33 & 0 \\
    (39.0\%) & (13.7\%) & (15.1\%) & (22.9\%) & (0.0\%) \\ 
    \bottomrule
    \end{tabular}
\end{table}

\subsection{Dataset}\label{subsec:results_dataset}

The xView3-SAR \cite{Paolo2022} dataset is a large and diverse collection of high-resolution SAR maritime images, including 754 publicly available images with an average size of 29,400 by 24,400 pixels. The scenes are C-band SAR amplitude images acquired by the European Space Agency's (ESA's) Sentinel-1 constellation; specifically the Interferometric Wide Swath mode Level-1 Ground Range Detected product, which includes both vertical-horizontal (VH) and vertical-vertical (VV) polarization bands for each image at a 10-meter pixel spacing. The scenes are supplied having undergone pre-processing, including orbit correction, removal of noise artifacts, radiometric calibration, terrain correction, and geocoding. In addition to the VH and VV SAR images, ancillary data are provided including wind speed, direction, quality, and land/ice mask information from Sentinel-1 Level-2 Ocean product, and bathymetry information from the General Bathymetric Chart of the Oceans (GEBCO, \cite{Weatherall2019}). These layers are provided in the xView3-SAR dataset co-registered with the respective SAR scenes and at a lower (500-meter) resolution than the SAR images. The scenes are distributed over four distinct geographical regions in European waters (Adriatic, Bay of Biscay, Iceland, and the North Sea) and one in West African waters (Gulf of Guinea). 

The xView3-SAR competition dataset provides labels for maritime object detections derived from two sources: the AIS database and manual expert reviewers. AIS is a radio transponder system that broadcasts ships' GNSS coordinates. The labels include geographic/pixel coordinates, class, bounding box coordinates and dimensions, length, distance from shore and confidence score. Class labels include non-vessels (fixed infrastructure such as wind farms, fish cages, platforms, and port towers), fishing vessels and non-fishing vessels. An example scene is shown in Fig. \ref{fig:SAR_satellite}a,b.

The dataset is divided into training (554 scenes), validation (50), and test sets (150). The training set contains automatically labeled (i.e. AIS) detections only, while the others contain both expert-reviewed AIS detection and manually-labeled detections. Consequently, the training dataset lacks manually annotated bounding boxes and is generally of lower quality, with occasional mismatches between bounding boxes and detected objects caused by the automated matching process. Additionally, some fixed infrastructure is missing ground-truth labels, and because many vessels do not broadcast AIS signals in near-shore areas, the training set contains approximately 100$\times$ fewer near-shore detections as a proportion of total detections than the validation set.

\subsection{Evaluation Metrics}\label{subsec:results_metrics}

To evaluate our models we use the test dataset, which contains 60,010 detections, and four detection and classification metrics from the xView3-SAR competition:
\begin{enumerate}
    \item Detection F1 score, evaluating identification of maritime objects across different sea states and geographical areas. 
	\item Near-shore F1 score, evaluating detection of maritime objects that are within two kilometers from shore (as defined by the zero-point in co-registered bathymetry data), in typically higher vessel density areas.
	\item Vessel Classification F1 score, evaluating performance at identifying whether a maritime object is a “vessel” or “fixed infrastructure” (non-vessel). This classification is only performed on manual labels with a high or medium confidence score, or that are identified via correlation with AIS.
	\item Fishing Classification F1 score, evaluating the model's performance in identifying fishing and non-fishing vessels. This classification is evaluated with AIS labels.
\end{enumerate}
This evaluation allows for direct comparison and benchmarking of our model to the models from the top-5 winning entries to the xView3-SAR competition \cite{Paolo2022}. The evaluation metrics for these models are reported in the top section of Table \ref{tab:yolo_performance} (blue symbols). The trained model weights and code for these models were also made available by competitors, as well as for the competition reference model (Faster R-CNN ResNet50). The competition also provides a vessel length estimation metric which is not used in this work, as our focus is on detection and classification performance. 

\subsection{Training Setup}\label{subsec:results_training}
Our training methodology employs two different datasets and two distinct training strategies to generate our final model. We utilize both the xView3-SAR training \textit{and} validation sets for training; the latter offers higher-quality annotations (see section C above), and has demonstrated superior detection performance when used as a training set in prior competition-winning solutions and in our own experiments. We therefore partitioned the xView3-SAR validation set into 40 scenes for training and 10 scenes for validation. These subsets are referred to as the \textit{training} and \textit{validation} sets throughout this work; the former is used for our initial exploratory analysis and for fine-tuning our final models. Additionally, we addressed the limitations of the automatically-generated labels in the original xView3-SAR training set by generating pseudo-labels and bounding boxes using our best-performing exploratory model (YOLOv8n-Ghost-P2-PIoU2 trained for 100 epochs). This enhanced dataset is referred to hereafter as the \textit{pre-training} set.

For initial model exploration (Table \ref{tab:yolo_performance}.A-D), we trained models on the \textit{training} set for 100 epochs using AdamW (batch size: 16, learning rate: $1.429\times10^{-3}$, momentum: 0.9, linear warm-up: 3 epochs). For the final model development (Table \ref{tab:yolo_performance}.E), we adopted a two-stage training procedure inspired by the transfer learning literature \cite{Xie2020, Zhong2019} and successful competition entries. This process involved first learning broadly transferable representations from a large, diverse, and potentially noisy dataset, followed by fine-tuning the model on a smaller dataset with clean labels for specialization. In the pre-training stage (\textbf{PT}), the model was trained for 100 epochs on the pseudo-labeled \textit{pre-training} dataset using AdamW (batch size: 32, learning rate: $7.14\times10^{-4}$, momentum: 0.9, linear warm-up: 3 epochs). This was followed by a fine-tuning stage (\textbf{FT}), in which the model was trained for another 100 epochs on the \textit{training} set using stochastic gradient descent with momentum (batch size: 16, learning rate: $5\times10^{-3}$, momentum: 0.9, no warm-up). Finally, we applied a threshold adjustment strategy (\textbf{TH}), in which the prediction threshold varies by predicted class label (non-vessel, non-fishing vessel, fishing vessel) and proximity to shore (within or outside 2 km). We optimized thresholds on the \textit{validation} set via grid search, selecting the configuration with the highest detection F1 score. We then fixed these thresholds and applied them unchanged to the held-out test set.

We trained all models using the Ultralytics codebase with PyTorch 2.4.1 and CUDA 12.1 on an NVIDIA A100 GPU (40 GB). The model input consisted of three channels; VV and VH SAR bands along with bathymetry data (e.g. see Fig. \ref{fig:SAR_satellite}f-h and Fig.~\ref{fig:yolov8_architecture}d), with scenes partitioned into 800 by 800 pixel chips using a tiling approach. SAR channel amplitudes were clipped to the range [-50, 20] dB, while bathymetry values were clipped to [-6000, 2000] meters, with each channel subsequently linearly normalized to the range [0, 255] for model input. We incorporated negative samples (background chips containing no detections) into the training data after identifying that this approach improved overall model detection performance by up to 2.9\% (see study in Table \ref{tab:model_ablations}.A). Consequently, we implemented a 20\% background ratio in the model training data, with background chips randomly sampled from the processed dataset. To enhance model generalization, we implemented standard image augmentations, including mosaic, random translations, scaling, left-right flips, image blurring, and hue-saturation-value adjustments.

\begin{table*}[t]
    \centering
    \caption{Performance of YOLOv8 detection models on the xView3-SAR Dataset. Subheadings indicate the specific model architecture and training variations explored. Architectures highlighted in bold represent the selected configurations carried forward to the final model ({\color[HTML]{15B01A}\raisebox{0.3ex}{\scriptsize{$\bigstar$}}}). xView3-SAR competition models are listed at the top, with the number of models in ensembles in parentheses. Colored symbols correspond to models illustrated in Fig. \ref{fig:xview3_models}. F1 scores for subsections A-E are provided as means and standard deviations over three random seeds.}
    \label{tab:yolo_performance}
    \renewcommand{\arraystretch}{.91}
    \begin{tabular}{@{}l@{\hspace{-14pt}}ccc cccc@{}}
    \toprule
    & & & & \multicolumn{4}{c}{\textbf{F1 Score}} \\
    \cmidrule(lr){5-8}
    \textbf{Model} & \textbf{Param (M)} & \textbf{Size (MB)} & \textbf{GFLOPs} & \textbf{Detection} & \textbf{Near-Shore} & \textbf{Vessel } & \textbf{Fishing} \\ 
    \midrule
    \multicolumn{8}{c}{\textit{xView3-SAR Competition }\cite{Paolo2022}}\\
    \mybluebullet CircleNet (12) & 322.5 & 1294.0 & 8766.9 & 0.755 & 0.525 & 0.942 & 0.834 \\
    {\color[HTML]{0066FF} $\blacksquare$} UNet (6)  & 432.5 & 1732.0 & 21079.1 & 0.752 & 0.473 & 0.938 & 0.803 \\
    {\color[HTML]{0066FF} \Large$\blackdiamond$} HRNet (15)  & 754.3 & 3029.0 & 11802.8 & 0.730 & 0.467 & 0.932 & 0.828 \\
    {\color[HTML]{0066FF} $\blacktriangledown$} Faster R-CNN & 89.1 & 249.0 & 5889.6 & 0.726 & 0.428 & 0.954 & 0.822 \\
    {\color[HTML]{0066FF} $\blacktriangle$} HRNet (12)  &  328.1 & 1314.0 & 426.5 & 0.725 & 0.461 & 0.940 & 0.797 \\
    \midrule

    \multicolumn{8}{c}{\textit{A) Model Size}} \\
    YOLOv8l & 43.7 & 175.4 & 165.2 & 0.607\tiny{$\pm$0.006} & 0.293\tiny{$\pm$0.020} & 0.895\tiny{$\pm$0.014} & 0.684\tiny{$\pm$0.021} \\
    YOLOv8m & 25.9 & 104.1 & 78.9 & 0.627\tiny{$\pm$0.014} & 0.364\tiny{$\pm$0.009} & 0.913\tiny{$\pm$0.004} & 0.679\tiny{$\pm$0.008} \\
    YOLOv8s & 11.2 & 45.1 & 28.6 & 0.622\tiny{$\pm$0.007} & 0.362\tiny{$\pm$0.006} & 0.915\tiny{$\pm$0.001} & 0.684\tiny{$\pm$0.010} \\
    \textbf{YOLOv8n} & 3.2 & 12.6 & 8.7 & 0.631\tiny{$\pm$0.008} & 0.367\tiny{$\pm$0.017} & 0.912\tiny{$\pm$0.006} & 0.693\tiny{$\pm$0.009} \\
    \midrule
    
    \multicolumn{8}{c}{\textit{B) Convolutional Block}} \\
    YOLOv8n-EfficientNet & 2.6 & 11.4 & 7.3 & 0.630\tiny{$\pm$0.003} & 0.367\tiny{$\pm$0.013} & 0.911\tiny{$\pm$0.003} & 0.708\tiny{$\pm$0.006}\\
    YOLOv8n-ShuffleNet & 1.8 & 7.6 & 5.3 & 0.626\tiny{$\pm$0.002} & 0.352\tiny{$\pm$0.002} & 0.904\tiny{$\pm$0.004} & 0.685\tiny{$\pm$0.002}\\
    YOLOv8n-FasterNet & 1.9 & 8.4 & 5.8 & 0.628\tiny{$\pm$0.005} & 0.359\tiny{$\pm$0.003} & 0.912\tiny{$\pm$0.003} & 0.704\tiny{$\pm$0.007}\\
    \textbf{YOLOv8n-Ghost} & 1.7 & 7.6 & 5.1 & 0.628\tiny{$\pm$0.002} & 0.361\tiny{$\pm$0.007} & 0.912\tiny{$\pm$0.001} & 0.708\tiny{$\pm$0.011} \\
    \midrule
    
    \multicolumn{8}{c}{\textit{C) P2 Feature Map}} \\
    YOLOv8n-Ghost-P2-3-4 & 1.1 & 4.6 & 8.3 & 0.664\tiny{$\pm$0.003} & 0.435\tiny{$\pm$0.005} & 0.919\tiny{$\pm$0.003} & 0.701\tiny{$\pm$0.003} \\
    \textbf{YOLOv8n-Ghost-P2} & 1.6 & 7.8 & 8.8 & 0.671\tiny{$\pm$0.002} & 0.437\tiny{$\pm$0.004} & 0.918\tiny{$\pm$0.003} & 0.705\tiny{$\pm$0.006} \\
    \midrule

    \multicolumn{8}{c}{\textit{D) Bounding Box Loss}} \\
    YOLOv8n-Ghost-P2-MPDIoU & 1.6 & 7.8 & 8.8 & 0.664\tiny{$\pm$0.002} & 0.437\tiny{$\pm$0.002} & 0.921\tiny{$\pm$0.006} & 0.718\tiny{$\pm$0.010} \\
    YOLOv8n-Ghost-P2-WIoU & 1.6 & 7.8 & 8.8 & 0.670\tiny{$\pm$0.004} & 0.420\tiny{$\pm$0.025} & 0.908\tiny{$\pm$0.014} & 0.697\tiny{$\pm$0.022}\\
    \textbf{YOLOv8n-Ghost-P2-PIoU2} & 1.6 & 7.8 & 8.8 & 0.685\tiny{$\pm$0.001} & 0.436\tiny{$\pm$0.004} & 0.917\tiny{$\pm$0.004} & 0.712\tiny{$\pm$0.008} \\
    \midrule

    \multicolumn{8}{c}{\textit{E) Two-Stage Training}} \\
    YOLOv8n (\textbf{PT})  & 3.2 & 12.6 & 8.7 & 0.621\tiny{$\pm$0.001} & 0.289\tiny{$\pm$0.004} & 0.873\tiny{$\pm$0.003} & 0.693\tiny{$\pm$0.008} \\  
    YOLOv8n (\textbf{FT})  & 3.2 & 12.6 & 8.7 & 0.632\tiny{$\pm$0.007} & 0.379\tiny{$\pm$0.011} & 0.928\tiny{$\pm$0.002} & 0.741\tiny{$\pm$0.013} \\  
    YOLOv8n (\textbf{FT}+\textbf{TH})  & 3.2 & 12.6 & 8.7 & 0.650\tiny{$\pm$0.002} & 0.385\tiny{$\pm$0.008} & 0.918\tiny{$\pm$0.003} & 0.737\tiny{$\pm$0.016} \\ 
    \addlinespace 
    YOLOv8s-Ghost-P2-PIoU (\textbf{PT})  & 5.3 & 22.6 & 22.3 & 0.692\tiny{$\pm$0.002} & 0.328\tiny{$\pm$0.007} & 0.869\tiny{$\pm$0.001} & 0.756\tiny{$\pm$0.002} \\  
    YOLOv8s-Ghost-P2-PIoU (\textbf{FT})  & 5.3 & 22.6 & 22.3 & 0.694\tiny{$\pm$0.003} & 0.448\tiny{$\pm$0.014} & 0.932\tiny{$\pm$0.002} & 0.769\tiny{$\pm$0.011}\\
    \myyellowbullet YOLOv8s-Ghost-P2-PIoU (\textbf{FT}+\textbf{TH})  & 5.3 & 22.6 & 22.3 & 0.703\tiny{$\pm$0.004} & 0.444\tiny{$\pm$0.010} & 0.928\tiny{$\pm$0.003} & 0.770\tiny{$\pm$0.013}\\
    \addlinespace 
    YOLOv8n-Ghost-P2-PIoU (\textbf{PT}) & 1.6 & 7.8 & 8.8 & 0.686\tiny{$\pm$0.001} & 0.324\tiny{$\pm$0.002} & 0.865\tiny{$\pm$0.003} & 0.743\tiny{$\pm$0.003} \\  
    YOLOv8n-Ghost-P2-PIoU (\textbf{FT}) & 1.6 & 7.8 & 8.8 & 0.691\tiny{$\pm$0.003} & 0.450\tiny{$\pm$0.003} & 0.929\tiny{$\pm$0.003} & 0.767\tiny{$\pm$0.006} \\  
    \mygreenbullet \textbf{YOLOv8n-Ghost-P2-PIoU} (\textbf{FT}+\textbf{TH})  & 1.6 & 7.8 & 8.8 & 0.704\tiny{$\pm$0.001} & 0.447\tiny{$\pm$0.007} & 0.925\tiny{$\pm$0.003} & 0.769\tiny{$\pm$0.008} \\ 
    \midrule

    \multicolumn{8}{c}{\textit{F) FPGA}} \\
    {\color[HTML]{FFC009} \raisebox{0.1em}{$\bigstar$}}YOLOv8s-Ghost-P2-PIoU2 & 5.3 & 5.1 & 37.2\raisebox{0.5ex}{\small *} & 0.697 & 0.438 & 0.931 & 0.785 \\
    {\color[HTML]{15B01A} \raisebox{0.1em}{$\bigstar$}}\textbf{YOLOv8n-Ghost-P2-PIoU2} & 1.6 & 1.5 & 14.6\raisebox{0.5ex}{\small *} & 0.704 & 0.445 & 0.921 & 0.765 \\
    \bottomrule
    \addlinespace[2pt]
    \multicolumn{8}{@{}l}{\raisebox{0.5ex}{\small *}INT8 operations (OPs) on FPGA DPU are reported instead of FLOPs due to model quantization.}
    \end{tabular}
\end{table*}

\section{Results}

We present the results of training the different YOLOv8 architectures in Section \ref{subsec:results_yolo}, the evaluation of the quantized model in Section \ref{subsec:results_compression}, and the performance of the deployed model running on FPGA hardware in Section \ref{subsec:results_fpga}.

\subsection{Model Modification Results}\label{subsec:results_yolo}

\begin{figure*}[t]
    \centering
    \includegraphics[width=\textwidth]{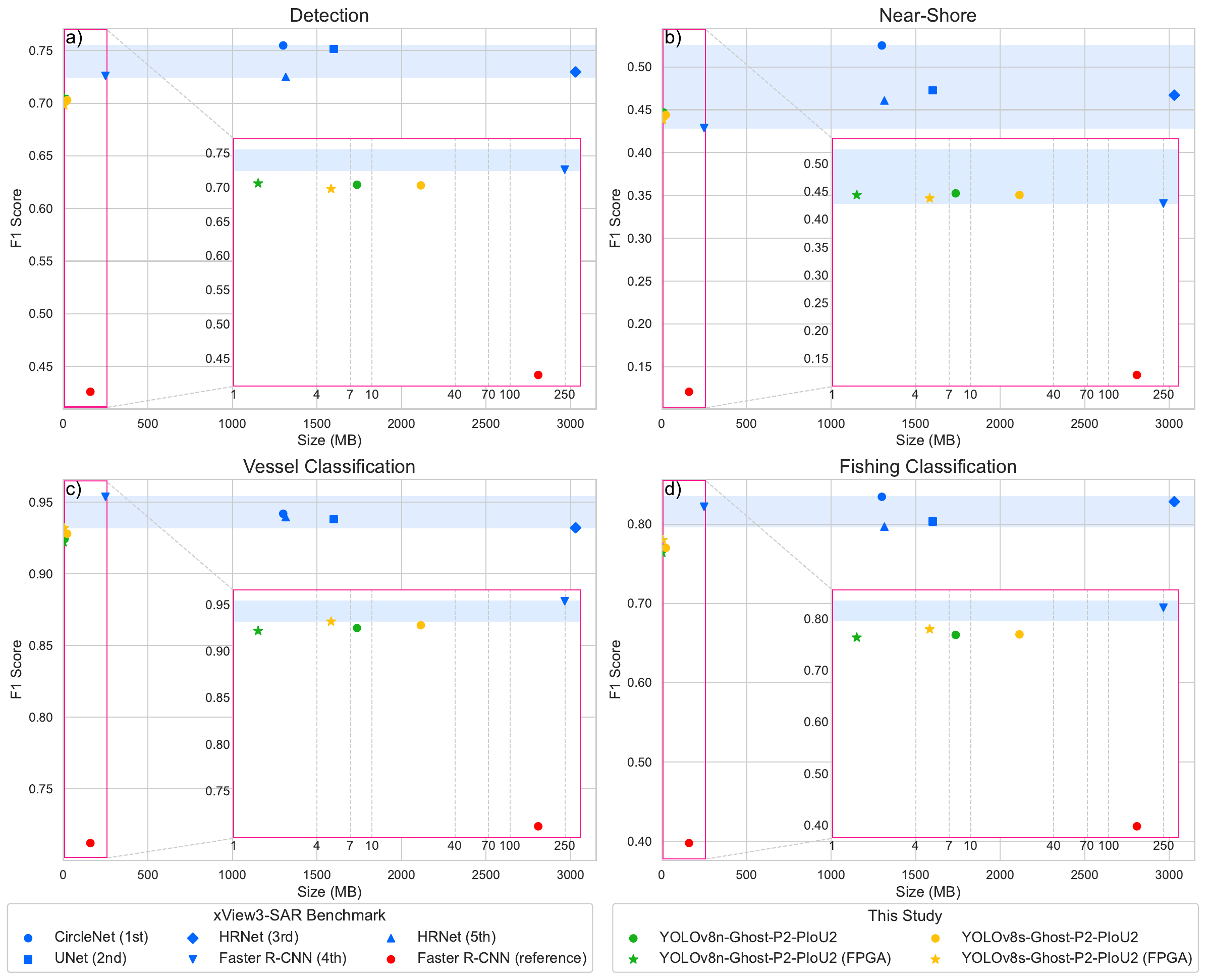}
    \caption{F1 Score vs. Model Size (MB) for Detection (a), Near-Shore Detection (b), Vessel Classification (c), and Fishing Vessel Classification (d) tasks. In each plot, xView3-SAR competition models are shown in blue, with the range of their F1 scores shown by the pale blue band. The competition reference model is shown by the red circle. Models from this study are shown in yellow (YOLOv8 \textit{small}) and green (YOLOv8 \textit{nano}), where floating point and quantized, FPGA-deployed models are denoted by circles and stars respectively. Inset plots in pink show a zoomed-in view of each corresponding plot between 0 and 250 MB on the x-axis (logarithmic scale).}
    \label{fig:xview3_models}
\end{figure*}

The model size, computational complexity and performance on the test dataset for a representative range of the various model architectures is presented in Table \ref{tab:yolo_performance}.
Our analysis of YOLOv8 model sizes (\textit{n}, \textit{s}, \textit{m} and \textit{l}) demonstrates that smaller architectures consistently outperform larger ones across most metrics (Table \ref{tab:yolo_performance}.A). The smallest model (\textit{nano}) shows improvements of 2.4\% in overall detection and 7.4\% in near-shore detection over the largest variant, has vessel classification accuracy within 0.3\% of larger models and outperforms larger models by up to 1.4\% in fishing classification. This inverse relationship between model complexity and performance (also seen by \cite{Lang2022}) may suggest that the maritime detection task benefits from simpler, more generalizable feature representations that avoid overfitting to training data complexities and AIS location errors.

Building upon the \textit{nano} baseline architecture, we explored efficiency-enhancing modifications that could maintain performance while reducing computational requirements. The integration of Ghost convolutions into the bottleneck layers (Table \ref{tab:yolo_performance}.B), delivers a 45\% reduction in model size and FLOPs while improving fishing classification performance by 1.5\% at the cost of a 0.3\% drop in overall detection. Alternative efficient convolutional blocks such as EfficientNet, ShuffleNet, and FasterNet \cite{Tan2019,Ma2018,Chen2023} achieved less substantial reductions in model size (ranging from 1.8-2.6M parameters compared to Ghost's 1.7M) while delivering comparable or slightly inferior performance, indicating that Ghost convolutions provide the most effective balance of efficiency and detection performance for this task.

The performance gains from architectural efficiency are further enhanced through multi-scale feature representation. Incorporating higher-resolution P2 feature maps into the \textit{nano} Ghost architecture improves near-shore detection performance by 7.6\% compared to the baseline YOLOv8n-Ghost model (Table \ref{tab:yolo_performance}.C). While adding the P2 features increases computational complexity by 72.5\%, the Ghost convolution's gains in efficiency offset this increase, keeping the final FLOPs equivalent to the baseline YOLOv8n model. Ablation studies confirm the importance of multi-scale features, as removing the P5 feature map (YOLOv8n-Ghost-P2-3-4) results in performance degradation of 0.7\% and 0.4\% in overall detection and fishing classification respectively compared to the P2 architecture. 

Building on these architectural improvements, we examined loss function variants specifically designed for small object detection. The PIoU2 provides consistent improvements of 1.4\% in overall detection and 0.7\% in fishing classification relative to the baseline CIoU (Table \ref{tab:yolo_performance}.D). In contrast, using MPDIoU loss results in improvements of 0.3\% and 1.3\% in classification but a decline in overall detection performance, while WIoU shows performance drops across all metrics compared to CIoU.

From the results of these architecture evaluation experiments, the YOLOv8n-Ghost-P2-PIoU2 model is selected as the best performing architecture; it combines the smallest base model size, the efficiency improvements of the Ghost blocks, the improved detection performance with the addition of the P2 feature map and the improved accuracy of the PIoU2 loss function. We further enhanced the model through the two-stage training approach as discussed in Section \ref{subsec:results_training} (Table \ref{tab:yolo_performance}.E). Pre-training on pseudo-labeled data (\textbf{PT}) consistently improves fishing classification performance due to increased sample diversity. This comes at the cost of reduced near-shore detection performance, likely caused by fewer AIS-broadcasting vessels in coastal areas, and therefore fewer ground-truth labels, but the subsequent fine-tuning stage on the \textit{training} set (\textbf{FT}) recovers these losses while maintaining classification gains. Overall this results in average improvements of 1.0\% in detection and 3.4\% in classification compared to single-stage training. Notably, following the same training procedure, the baseline unmodified YOLOv8n model trails our final model by 6.5\% and 1.3\% in detection and classification respectively. Finally, with the threshold adjustment strategy (\textbf{TH}) we are able to increase overall detection performance by 1.3\% on our YOLOv8n-Ghost-P2-PIoU2 model, with a 0.4\% average drop in the other metrics.

\begin{figure*}[t]
    \centering
    \includegraphics[width=.95\textwidth]{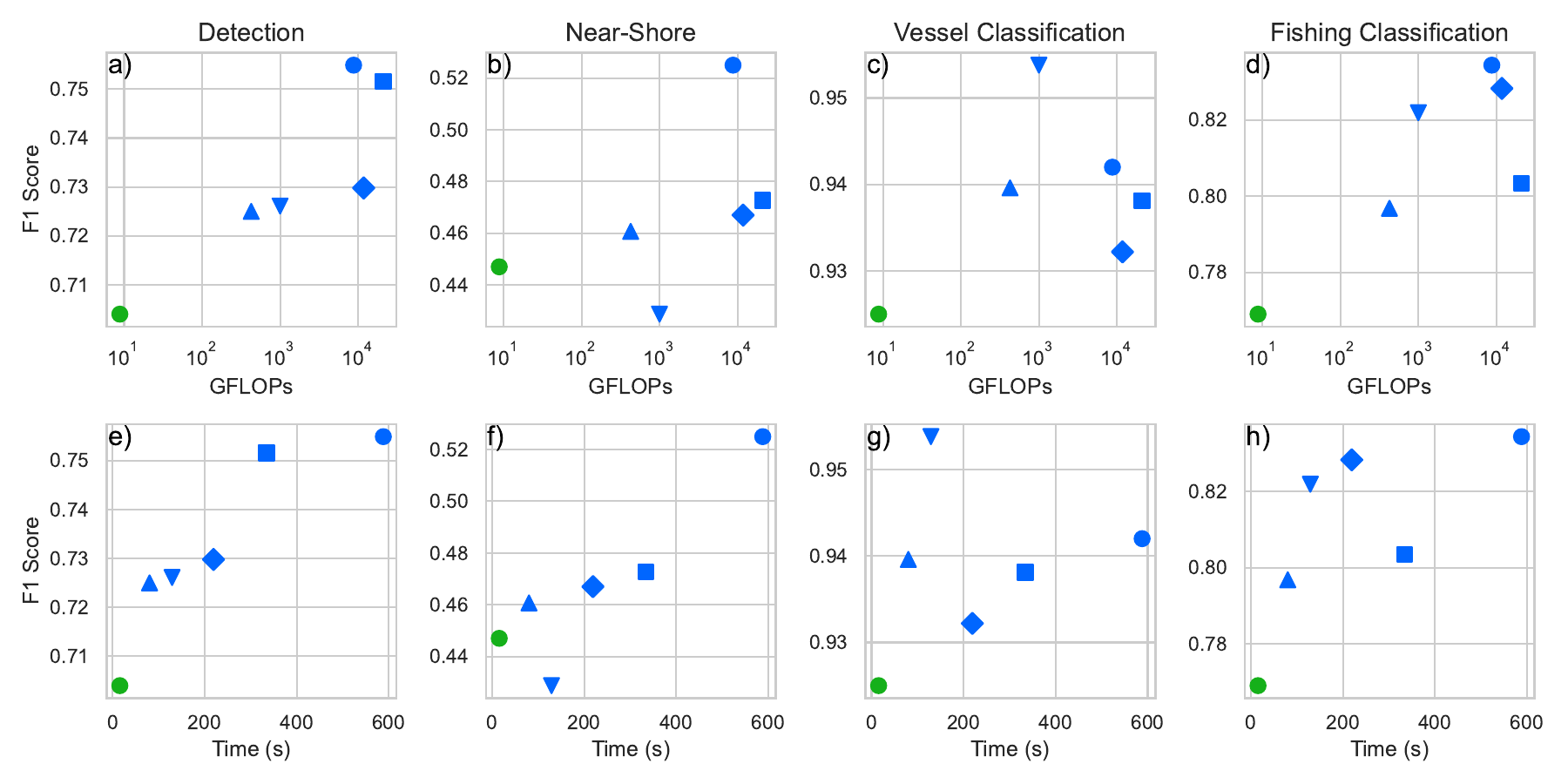}
    \caption{F1 Score vs. Model complexity (a-d; GFLOPs, logarithmic scale) and vs. Average inference time per scene (e-h; s, over three scenes) for Detection, Near-Shore Detection, Vessel and Fishing Vessel Classification. xView3-SAR competition top-5 models shown in blue and our YOLOv8n-Ghost-P2-PIoU2 model in green (see legend in Fig.~\ref{fig:xview3_models}).}
    \label{fig:time_perf_scatter}  
\end{figure*}

Comparing the performance of our best models to the top-5 winning entries of the xView3-SAR competition (see Fig. \ref{fig:xview3_models}), our model nears the competition models with between 98.2\% and 99.8\% fewer parameters. Our detection performance is within 2.1\% of the top-5 winning entries and outperforms one model in near-shore detection by 1.9\%. Our model is only 0.7\% and 2.8\% behind the top-5 in vessel and fishing classification respectively.
We also investigated several additional modifications, following methods from the literature, which did not show improvements for our implementation. Modifying the model's FPN to AFPN or GFPN showed a performance drop across the different metrics, with slight model size increases. Similarly, incorporating attention modules (Convolutional Block Attention Module \cite{Woo2018}, CA, EMA) at the point where feature maps are fed into the FPN improved fishing classification performance by an average of 2.8\%, though this came at the cost of reduced overall detection performance. 
Our proposed model is substantially more computationally efficient than existing methods (8.8 GFLOPs rather than 426.5--21079.1). This corresponds to a 98.0\% to 99.9\% reduction in computational complexity, without performance loss beyond observed efficiency-performance trends (Fig. \ref{fig:time_perf_scatter}.a-d). This efficiency is reflected by inference speed on an NVIDIA A100 GPU, where our model processes a scene in 15.4 seconds on average, achieving a 5.2$\times$ to 38.2$\times$ speed-up over alternatives (80 to 588 seconds; Fig. \ref{fig:time_perf_scatter}.e-h). The magnitude of the speed-up is less pronounced than FLOPs reduction, as the GPU's architecture is highly effective at parallelizing the workloads of the more computationally intensive models, and our model is not optimized for GPU.

\begin{table}[h]
    \centering
    \caption{Impact of the Background Ratio and Auxiliary Channel on YOLOv8n-Ghost-P2-PIoU2 performance.}
    \label{tab:model_ablations}
    \renewcommand{\arraystretch}{.95}
    \begin{tabular}{@{}lcccc@{}}
    \toprule
    & \multicolumn{4}{c}{\textbf{F1 Score}} \\
    \cmidrule(lr){2-5}    
    \textbf{Metric} & \textbf{Detection} & \textbf{Near-Shore} & \textbf{Vessel} & \textbf{Fishing} \\ 
    \midrule
    \multicolumn{5}{c}{\textit{A) Background Ratio}} \\
    40 & 0.680\tiny{$\pm$0.015} & 0.433\tiny{$\pm$0.005} & 0.917\tiny{$\pm$0.004} & 0.711\tiny{$\pm$0.011} \\ 
    30 & 0.684\tiny{$\pm$0.001} & 0.436\tiny{$\pm$0.004} & 0.917\tiny{$\pm$0.002} & 0.706\tiny{$\pm$0.005} \\ 
    20 & 0.685\tiny{$\pm$0.001} & 0.436\tiny{$\pm$0.004} & 0.917\tiny{$\pm$0.004} & 0.712\tiny{$\pm$0.008} \\ 
    10 & 0.672\tiny{$\pm$0.007} & 0.417\tiny{$\pm$0.012} & 0.914\tiny{$\pm$0.004} & 0.705\tiny{$\pm$0.022} \\ 
    0  & 0.656\tiny{$\pm$0.016} & 0.416\tiny{$\pm$0.017} & 0.915\tiny{$\pm$0.003} & 0.701\tiny{$\pm$0.018} \\ 
    \midrule
    \multicolumn{5}{c}{\textit{B) Auxiliary Channel}} \\
    Bath & 0.685\tiny{$\pm$0.001} & 0.436\tiny{$\pm$0.004} & 0.917\tiny{$\pm$0.004} & 0.712\tiny{$\pm$0.008} \\
    $\emptyset$ & 0.690\tiny{$\pm$0.007} & 0.423\tiny{$\pm$0.017} & 0.907\tiny{$\pm$0.013} & 0.694\tiny{$\pm$0.014} \\
    \bottomrule
    \end{tabular}
\end{table}

\subsection{Quantization Results}\label{subsec:results_compression}

We employ quantization to reduce model size and computational complexity, as well as to meet the constraints of the target FPGA hardware. As the SiLU activation is not supported by the framework, we replace it with Hard Swish and finetune (\textbf{FT}) the model, with negligible deterioration to the float model (Table \ref{tab:quant_study}.A). We evaluate quantized models on the test dataset on GPU, and Table \ref{tab:yolo_quantization} compares performance across our YOLOv8 variants, all trained with two-stage training and modified activation.

\subsubsection{Post-Training Quantization (PTQ)}
We calibrated our models on a 1,000 chip subset of the \textit{validation} set. Smaller (100) and larger (10,000) calibration sets were also tested, but these changes did not result in significant performance differences.
With PTQ, models show moderate performance degradation compared to their floating-point counterparts, with a minor overall detection drop of 0.4\% and a decrease of an average of 3.4\% in near-shore and classification metrics. The baseline YOLOv8n model experiences the largest performance drop, with an average performance drop of 4.1\% across all metrics, compared to 1.7\% and 2.2\% for our modified \textit{small} and \textit{nano} models respectively. These results indicate that quantization exacerbated the limitations of the less performant baseline architecture, while our modified architectures are more robust to quantization. Additionally, the increased performance drop of the smallest model (\textit{nano}) can be attributed to the reduced amount of redundancy with fewer parameters, which leads to a more pronounced impact. 

\begin{table}[h]
    \centering
    \caption{Performance of YOLOv8 model variants with INT8 PTQ and QAT on GPU, and QAT on FPGA.}
    \label{tab:yolo_quantization}
    \renewcommand{\arraystretch}{.95}
    \begin{tabular}{@{}l@{\hspace{-1pt}}cccc@{}}
    \toprule
    & \multicolumn{4}{c}{\textbf{F1 Score}} \\
    \cmidrule(lr){2-5}    
    \textbf{Model} & \textbf{Detection} & \textbf{Near-Shore} & \textbf{Vessel } & \textbf{Fishing} \\ 
    \midrule
    
    \multicolumn{5}{c}{\textit{A) YOLOv8n (\textbf{PT}+\textbf{FT}+\textbf{TH})}} \\
    Float &  0.650\tiny{$\pm$0.003} & 0.384\tiny{$\pm$0.003} & 0.914\tiny{$\pm$0.003} & 0.731\tiny{$\pm$0.007} \\
    PTQ  & 0.644\tiny{$\pm$0.003} & 0.327\tiny{$\pm$0.012} & 0.872\tiny{$\pm$0.016} & 0.671\tiny{$\pm$0.012} \\
    QAT  & 0.646\tiny{$\pm$0.004} & 0.377\tiny{$\pm$0.011}  & 0.915\tiny{$\pm$0.002} & 0.725\tiny{$\pm$0.018} \\
    FPGA & 0.647 & 0.386 & 0.914 & 0.730 \\
    \midrule

    \multicolumn{5}{c}{\textit{B) YOLOv8s-Ghost-P2-PIoU2 (\textbf{PT}+\textbf{FT}+\textbf{TH})}} \\
    \myyellowbullet Float & 0.701\tiny{$\pm$0.001} & 0.442\tiny{$\pm$0.003} & 0.920\tiny{$\pm$0.018} & 0.774\tiny{$\pm$0.013}\\
    PTQ & 0.696\tiny{$\pm$0.005} & 0.420\tiny{$\pm$0.015} & 0.903\tiny{$\pm$0.010} & 0.752\tiny{$\pm$0.013}\\
    QAT & 0.699\tiny{$\pm$0.001} & 0.444\tiny{$\pm$0.007} & 0.932\tiny{$\pm$0.001} & 0.779\tiny{$\pm$0.001}\\
    {\color[HTML]{FFC009}\raisebox{0.1em}{$\bigstar$}}FPGA & 0.697 & 0.438 & 0.931 & 0.785 \\
    \midrule
    
    \multicolumn{5}{c}{\textit{C) YOLOv8n-Ghost-P2-PIoU2 (\textbf{PT}+\textbf{FT}+\textbf{TH})}} \\
    \mygreenbullet Float & 0.699\tiny{$\pm$0.004} & 0.444\tiny{$\pm$0.004} & 0.923\tiny{$\pm$0.002} & 0.769\tiny{$\pm$0.006} \\
    PTQ  & 0.697\tiny{$\pm$0.005} & 0.419\tiny{$\pm$0.009} & 0.892\tiny{$\pm$0.004} & 0.740\tiny{$\pm$0.004} \\
    QAT  & 0.701\tiny{$\pm$0.003} & 0.445\tiny{$\pm$0.001} & 0.927\tiny{$\pm$0.005} & 0.771\tiny{$\pm$0.009} \\
    {\color[HTML]{15B01A} \raisebox{0.1em}{$\bigstar$}}FPGA & 0.704 & 0.445 & 0.921 & 0.765 \\
    \bottomrule
    \end{tabular}
\end{table}

\subsubsection{Quantization-Aware Training (QAT)}
\begin{figure}[h]
    \centering
    \includegraphics[width=.5\textwidth]{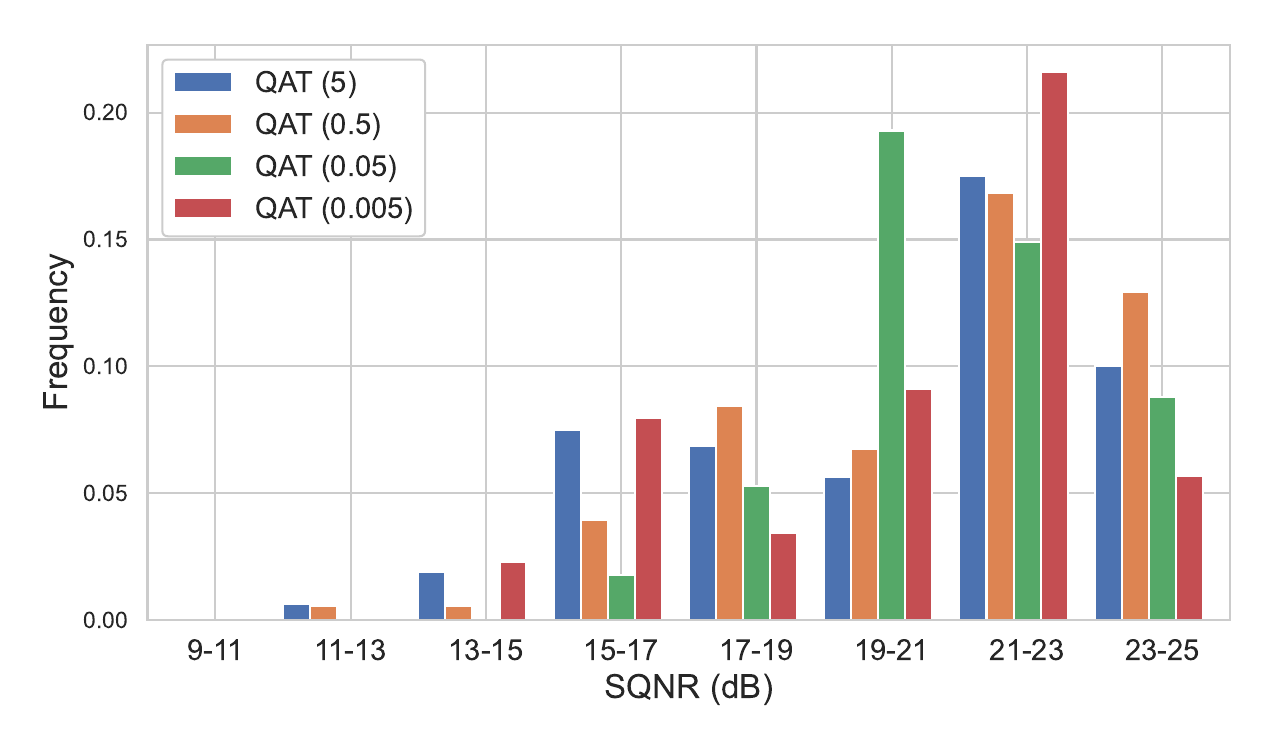}
    \caption{Histogram of per-subgraph SQNR for different QAT learning rates for YOLOv8n-Ghost-P2-PIoU2.}\label{fig:quant_sqnr}  
\end{figure}

To minimize the performance drop observed with PTQ, we modify the second training stage (\textbf{FT}) with QAT. We use the same training parameters as the fine-tuning stage detailed in Section \ref{subsec:results_training}, with the addition of a quantizer parameter learning rate. We study the impact of the quantizer learning rate on the model performance (Table \ref{tab:quant_study}) and signal-to-quantization-noise ratio (SQNR; Fig. \ref{fig:quant_sqnr}) of the model subgraphs (partitions of the computational graph). The SQNR provides a measure of the fidelity of the quantized representation, where higher values correspond to reduced distortion due to quantization and is defined as SQNR$=10\cdot\log_{10}(\frac{P_{\text{signal}}}{P_{\text{noise}}})$. We find that higher learning rates (5 and 0.5) produce a broader distribution, with higher SQNR modes (23 and 24 dB) with a longer tail towards lower SQNR values. Lower learning rates (0.05 and 0.005) produce a more concentrated distribution, with a lower SQNR mode (20 dB) and fewer low SQNR subgraphs. However, the detection and classification performance difference is minimal across the different values, and we therefore opt for a learning rate of 0.5 in our solution. QAT implementation introduces minimal performance degradation across most metrics (average -0.4\% for detection F1), with our modified architecture models showing slight improvement in classification metrics (e.g. +0.9\% for YOLOv8s-Ghost-P2-PIoU2 QAT). The baseline architecture shows the highest average performance drop of 0.7\%, with the \textit{nano} modified architectures showing a loss of 0.1\% and the \textit{small} a 0.2\% improvement overall. Nevertheless, the YOLOv8n-Ghost-P2-PIoU2 model shows the highest detection performance overall, with a 0.701 and 0.445 F1 score for detection and near-shore detection respectively, outperforming the baseline model and the \textit{small} model by 5.5\% and 0.2\% respectively. These results indicate that with QAT, most of the performance drop observed with PTQ can be recovered, and confirms that our two-stage training approach effectively improves model performance.

\begin{table}[h]
    \centering
    \caption{Impact of Activation and Quantizer Learning Rate on YOLOv8n-Ghost-P2-PIoU2 performance.}
    \label{tab:quant_study}
    \renewcommand{\arraystretch}{.95}
    \begin{tabular}{@{}l@{\hspace{8.5pt}}cccc@{}}
    \toprule
    & \multicolumn{4}{c}{\textbf{F1 Score}} \\
    \cmidrule(lr){2-5}    
    \textbf{Metric} & \textbf{Detection} & \textbf{Near-Shore} & \textbf{Vessel} & \textbf{Fishing} \\ 
    \midrule
    \multicolumn{5}{c}{\textit{A) Activation Function (\textbf{FT})}} \\
    SiLU & 0.691\tiny{$\pm$0.003} & 0.450\tiny{$\pm$0.003} & 0.929\tiny{$\pm$0.003} & 0.767\tiny{$\pm$0.006} \\ 
    HSwish & 0.686\tiny{$\pm$0.005} & 0.446\tiny{$\pm$0.003} & 0.928\tiny{$\pm$0.003} & 0.766\tiny{$\pm$0.006} \\ 
    \midrule
    \multicolumn{5}{c}{\textit{B) Quantizer Learning Rate (QAT \textbf{FT})}} \\
    5   & 0.690\tiny{$\pm$0.004} & 0.443\tiny{$\pm$0.002} & 0.929\tiny{$\pm$0.002} & 0.765\tiny{$\pm$0.007} \\
    0.5   & 0.693\tiny{$\pm$0.005} & 0.443\tiny{$\pm$0.001} & 0.930\tiny{$\pm$0.001} & 0.763\tiny{$\pm$0.003} \\
    0.05  & 0.686\tiny{$\pm$0.010} & 0.439\tiny{$\pm$0.006} & 0.931\tiny{$\pm$0.003} & 0.765\tiny{$\pm$0.002} \\
    0.005 & 0.688\tiny{$\pm$0.003} & 0.444\tiny{$\pm$0.013} & 0.929\tiny{$\pm$0.008} & 0.765\tiny{$\pm$0.008} \\
    \bottomrule
    \end{tabular}
\end{table}

\subsection{FPGA Deployment Results}\label{subsec:results_fpga}

\begin{figure}
    \centering
    \includegraphics[width=.48\textwidth]{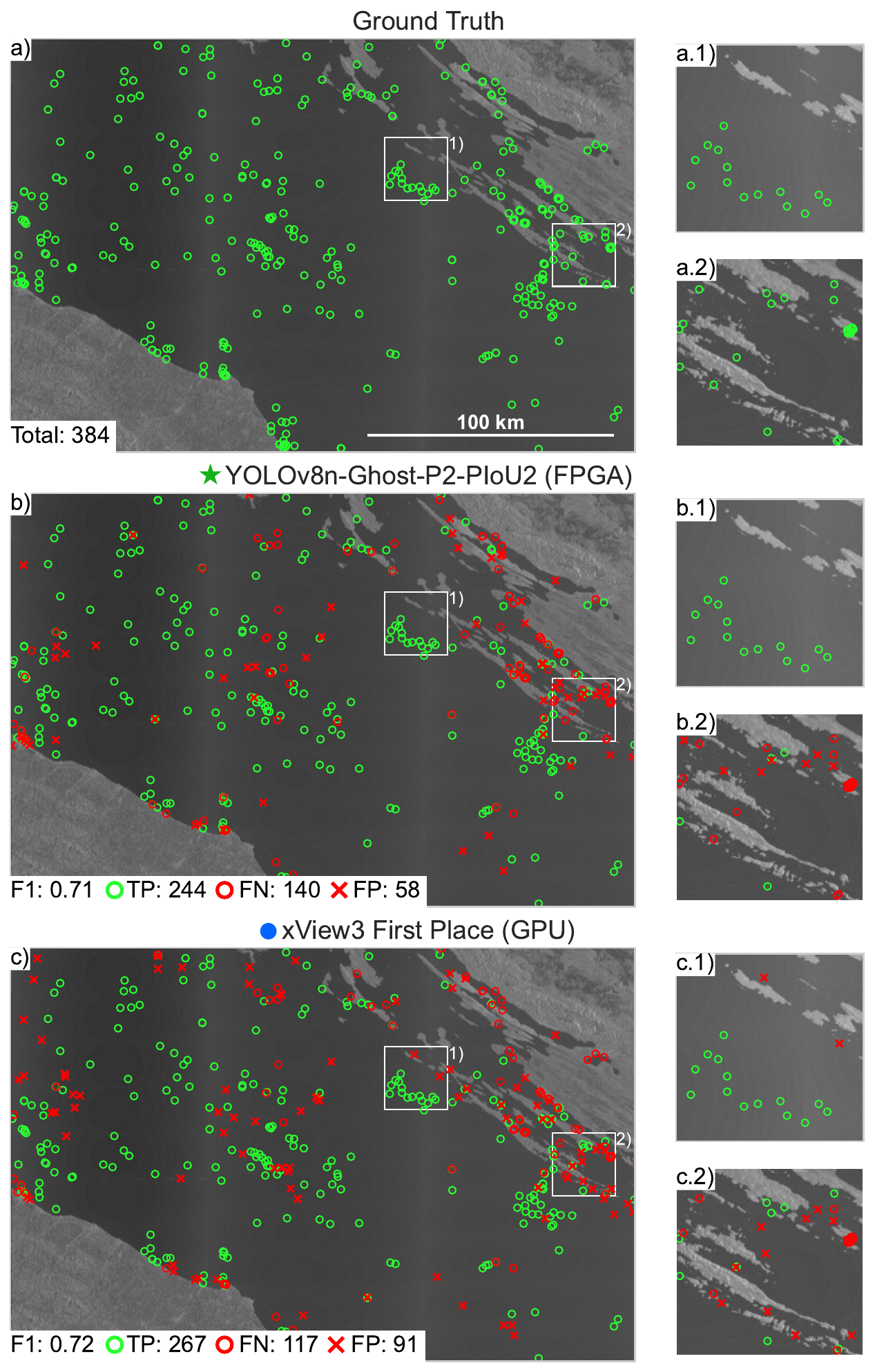}
    \caption{Example scene from xView3-SAR comparing (a) ground truth, (b) predictions from our YOLOv8n-Ghost-P2-PIoU2 model on FPGA, and (c) the first-place competition entry. Insets highlight areas of interest.}  
    \label{fig:qualitative_results}
\end{figure}

The performance of the deployed compiled QAT models on the Kria KV260 board is shown in Table \ref{tab:yolo_quantization} and matches the expected performance based on the GPU results, with little to no performance degradation observed. For our final model, we also visually illustrate the quality of the results with respect to a state-of-the-art GPU-based model. From simple visual inspection of these qualitative results (Fig. \ref{fig:qualitative_results}), our FPGA-deployed model shows similar levels of performance to the xView3-SAR first-place model. We note that the winning model shows slightly higher performance overall by missing fewer detections (false negatives), at the cost of a larger number of false positives. The winning model also shows slightly higher detection performance in near-shore areas, where our model misses more detections (contrast Fig. \ref{fig:qualitative_results}.b.2 vs c.2). However, it is important to note that our model is only 0.15\% of the size of the winning model (Table \ref{tab:yolo_performance}).

The power usage, efficiency, throughput and external memory average bandwidth of the YOLOv8 model variants implemented on the Kria KV260 development board are presented in Table \ref{tab:fpga_results}, where we provide performance of both the `Full' pipeline implementations (including pre and post-processing) and of inference alone (`DPU'). SoC power measurements were collected using xlnx\_platformstats monitoring software with a 30-second sampling window and samples taken every 0.01 seconds (100 Hz sampling rate). Average power consumption was calculated over the full inference period, while peak power represents the highest reading recorded during operation. When idle, the SoC has a baseline consumption of 4.1 watts.

\begin{table}[h]
    \centering
     \caption{Power, throughput (Tput), full implementation average power efficiency (Full Eff), and average memory bandwidth (BW) of YOLOv8 variants inference on Kria KV260 vs. thread count (1, 2, 4 and 8).}
     \label{tab:fpga_results}
     \renewcommand{\arraystretch}{.9}
    \begin{tabular}{@{}cccccccc@{}}
    \toprule
    \textbf{} & \multicolumn{2}{c}{\textbf{Power (W)}} & \multicolumn{2}{c}{\textbf{Tput (FPS)}} & \textbf{Full Eff} & \multicolumn{2}{c}{\textbf{BW (MB/s)}} \\
    \cmidrule(lr){2-3} \cmidrule(lr){4-5} \cmidrule(lr){7-8}
    \textbf{(\#)} &  \textbf{Avg} &  \textbf{Peak} & \textbf{Full} & \textbf{DPU} & \textbf{(FPS/W)} & \textbf{Full} & \textbf{DPU} \\
    \midrule
    \multicolumn{8}{c}{\textit{A) YOLOv8n}} \\
   1 & 5.4 & 8.9 & 6.9 & 24.7 & 1.3 & 4652 & 4650 \\
   2 & 6.4 & 9.4 & 12.8 & 29.8 & 2.0 & 4392 & 4570 \\
   4 & 7.6 & 9.8 & 19.8 & 29.8 & 2.6 & 3664 & 4571 \\
   8 & 7.8 & 9.8 & 21.6 & 29.8 & 2.8 & 3442 & 4570 \\
   \midrule
   \multicolumn{8}{c}{\textit{B) YOLOv8n-Ghost}} \\
   1 & 5.1 & 8.6 & 7.2 & 27.0 & 1.4 & 4375 & 4374 \\
   2 & 6.0 & 9.2 & 13.4 & 33.6 & 2.2 & 4151 & 4319 \\
   4 & 7.2 & 9.4 & 20.9 & 33.6 & 2.9 & 3606 & 4319 \\
   8 & 7.3 & 9.7 & 22.9 & 33.6 & 3.1 & 3372 & 4318 \\
   \midrule
    \multicolumn{8}{c}{\textit{C) YOLOv8n-Ghost-P2}} \\
   1 & 4.9 & 9.5 & 3.6 & 13.3 & 0.7 & 4701 & 4692 \\
   2 & 5.7 & 10.0 & 7.0 & 20.5 & 1.2 & 4564 & 4599 \\
   4 & 6.9 & 10.5 & 11.8 & 20.5 & 1.7 & 4223 & 4601 \\
   8 & 7.0 & 10.8 & 13.4 & 20.5 & 1.9 & 4033 & 4603 \\
   \midrule
    \multicolumn{8}{c}{\textit{D) YOLOv8s-Ghost-P2}} \\
   1 & 5.3 & 9.8 & 3.2 & 9.3 & 0.6 & 5259 & 5259 \\
   2 & 6.5 & 10.2 & 6.2 & 12.2 & 1.0 & 5196 & 5203 \\
   4 & 8.1 & 10.4 & 10.4 & 12.2 & 1.3 & 4710 & 5202 \\
   8 & 8.1 & 10.4 & 10.5 & 12.2 & 1.3 & 4692 & 5203 \\
    \bottomrule
\end{tabular}
\end{table}

Architectural modifications to the YOLOv8n model yield distinct trade-offs on the FPGA platform. The integration of Ghost blocks provides a 10.5\% increase in DPU-only throughput, while reducing the end-to-end power consumption and average DPU memory-bandwidth by 5.9\%, confirming the efficiency improvement of our architecture on hardware. In a single-threaded environment, the end-to-end throughput across all models is on average 3.5 times lower than the DPU-only inference capabilities, highlighting a severe CPU bottleneck. Increasing the CPU thread count mitigates this limitation, increasing throughput on average by a factor of 3.2 at the cost of increased average and peak power consumption, but performance gains plateau as the system becomes DPU-bound. For YOLOv8n-Ghost, throughput scales to 22.9 FPS with eight threads, approaching its DPU ceiling of 33.6 FPS. The P2 architecture models are more computationally expensive, with the 14.6 GOPs workload (Table \ref{tab:yolo_performance}) of the YOLOv8n-Ghost-P2 model nearly double the YOLOv8n-Ghost's workload (7.9 GOPs). Whilst the YOLOv8n-Ghost-P2 model has a similar workload to that of the baseline YOLOv8n model (12.7 GOPs), its DPU-only throughput is 46\% and 31\% lower with one and two threads respectively. This discrepancy highlights a DPU-memory bottleneck, which is caused by its 52\% higher external memory load size from additional high-resolution feature maps. However, the P2 models still shows the largest end-to-end throughput improvements from one to eight threads, with the \textit{nano} and \textit{small} models improving by a factor of 3.7 and 3.3 respectively. Analysis of memory bandwidth also shows that as thread count increases, the full-pipeline average bandwidth decreases (e.g. from 4375 to 3372 MB/s for YOLOv8n-Ghost), while the DPU-only bandwidth remains constant. This indicates significant memory bus contention between the parallel CPU threads and the DPU, which throttles the DPU's data access. Similarly, throughput of DPU-only implementations does not increase past two threads, suggesting that threadcount could be used to balance performance against deployment power constraints.

\begin{figure}[t]
    \centering
    \includegraphics[width=.48\textwidth]{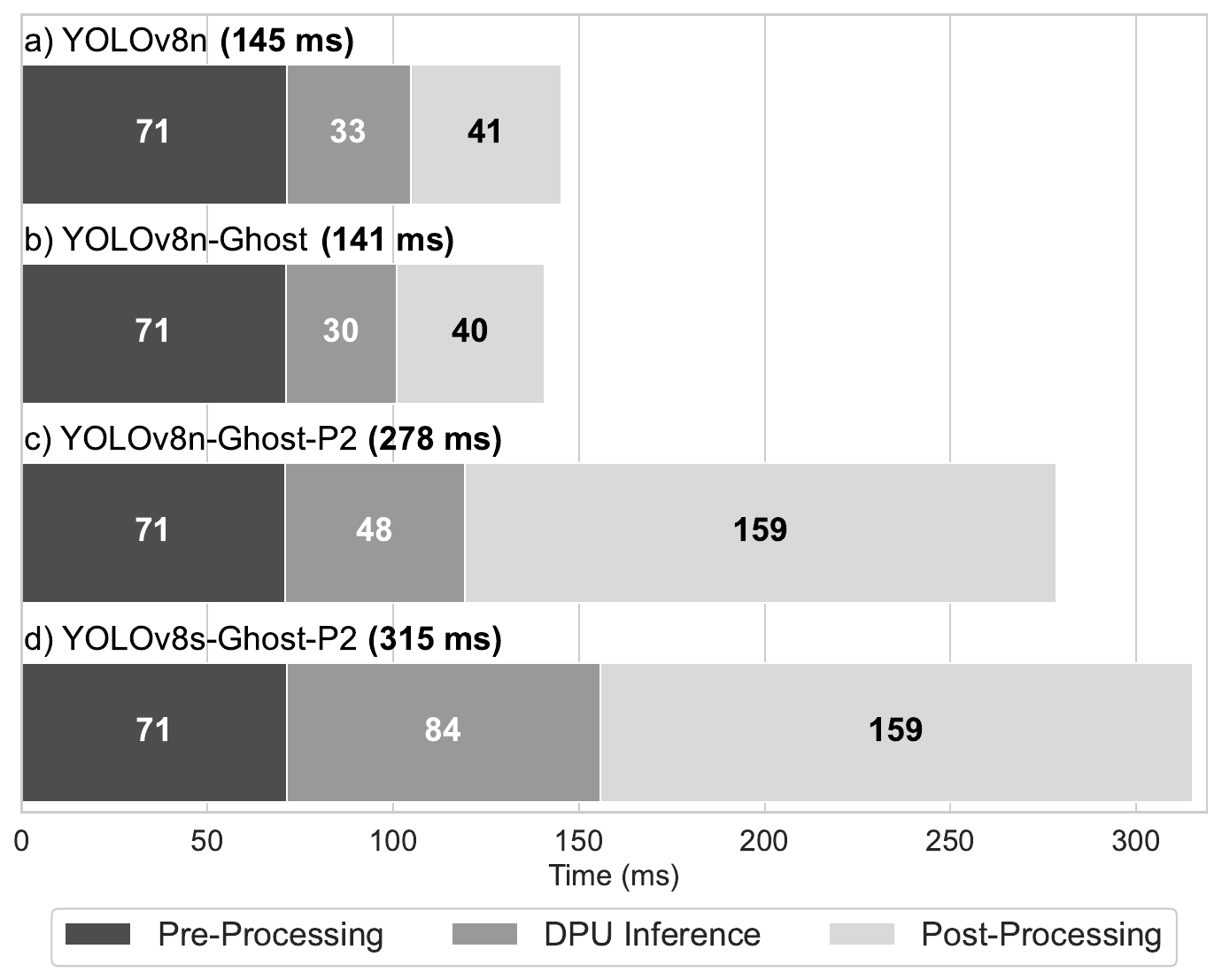}
    \caption{Processing time breakdown of YOLOv8 variants on Kria KV260 (average from 100 inference runs).}
    \label{fig:inference_breakdown}
\end{figure}

A breakdown of the inference time for different YOLOv8 model architectures is shown in Fig. \ref{fig:inference_breakdown}. Image loading and pre-processing requires on average 71 ms, while DPU inference times vary by architecture. The baseline \textit{nano} YOLOv8 model has an average inference time of 33 ms. The Ghost variant achieves a 9\% reduction in latency, with an average of 30 ms, attributed to a 41\% decrease in computational complexity. However, the speed-up is partially offset by increased memory access overhead due to higher read/write operations. The P2 variant is 60\% slower (48 ms) due to its additional layers and larger feature maps. Finally, the larger \textit{small} Ghost P2 model is the slowest at 84 ms. The post-processing step takes on average 41 ms for the baseline architectures and 159 ms for the P2 variants. The increased computation time results from the P2 feature map generating 40,000 additional predictions compared to the baseline configuration. Post-processing consists of two stages: a computationally intensive decoding step that transforms raw outputs into bounding box coordinates, class labels and confidence scores; and Non-Maximum Suppression (NMS) filtering (only takes 91~$\mu$s on average). 

\section{Discussion}

In this study, we have made a key contribution towards on-satellite deployment of ML models for SAR vessel detection. We now discuss further steps towards real-world usage of our model, in terms of: A) directions for further improvement of model performance and efficiency; B) technical considerations for practical on-satellite deployment. 

\subsection{Potential further performance and efficiency improvement}

The residual performance gap between our FPGA-deployed model and larger GPU-based models is mainly in near-shore detection (0.447 vs. 0.428--0.525 F1 score of the xView3-SAR top-5 entries) and fishing classification (0.769 vs. 0.797--0.834 F1 score), and is likely partly due to our focus on inference efficiency. All five of our benchmark models employ model ensembles, test-time augmentation (TTA), or both during inference, which improve performance but which we choose not to implement here due to increased computational cost.
The performance difference may also reflect a fundamental limit of the simple single-stage detection architecture for this challenging task, as several winning xView3-SAR competitors used architectures such as UNet or HRNet, which are large but excel at preserving the fine-grained spatial details crucial for distinguishing small vessels from complex coastal clutter. Vision Transformers may be a viable lightweight alternative to CNNs for this task, and have shown promise both for SAR vessel detection \cite{Ahmed2024,Gupta2024,Zhang2025Preprint}, and for maintaining performance at low bit-width, e.g. binary and ternary compression \cite{Xu2022TerViT,He2023}. Alternatively, further performance improvements may be possible by combining our model with statistical approaches, such as CFAR \cite{Wen2024}, or through leveraging additional auxiliary information (e.g. coastline, static infrastructure or areas of interest) or historical imagery.

Despite the overall speed of our current analysis pipeline, there are also opportunities to further improve our model's speed of analysis and efficiency.
There is a notable computational bottleneck in both pre- and post-processing operations, with these steps consuming 83\% of the total inference time (71 ms and 159 ms respectively, in comparison to just 48 ms for inference itself, see Fig. \ref{fig:inference_breakdown}). 
The post-processing overhead is the largest, and arises from additional tens of thousands of outputs generated by the high-resolution P2 feature map, which require decoding into bounding box predictions. Simple modifications to our decoding code already achieved a 28$\times$ efficiency improvement over the baseline Vitis AI implementation, reducing processing time from 4400 ms to 159 ms, and we consider two additional approaches to reduce this further. We first test a hybrid strategy where the P2 model variant is only used on the more challenging near-shore areas of each scene (6\% of the test dataset, identified using bathymetry, but could also be identified from SAR metadata and an onboard look-up table), while the baseline YOLOv8n model processes the remaining regions. This approach significantly reduces the average total inference time, from 278 ms to 153 ms per chip across the test set, but also leads to a significant performance drop of 4.1\% and 4.3\% in overall detection and fishing classification respectively compared to our preferred model.
Alternatively, we also conducted preliminary experiments to offload the decoding step to the PL using Vitis HLS for operation parallelization. The synthesis results show a moderate latency reduction to 133 ms, and importantly, this approach also decreases PS–PL data transfer. This is a promising avenue for real-world deployment, running in parallel to the DPU by leveraging the unused resources of the FPGA's PL (see Table \ref{tab:fpga_resources} for utilization). We leave deployment on hardware to future work. 
Pre-processing also likely carries unnecessary overhead due to ingestion of 32-bit floating point TIFF files and then conversion to INT8 images; this could be reduced by storing image chips in lower precision, and by implementing image normalization operations directly on the FPGA as per the post-processing. 

Additional efficiency gains could come from systematic optimization of the DPU configuration, e.g. by varying parallelism levels, core counts, and operating frequencies; this is beyond current scope but  2$\times$ improvement in power efficiency has previously been demonstrated on identical hardware \cite{Hamanaka2023}. Finally, we could investigate quantizing our model to lower bit-widths (e.g. 4-bit \cite{Yan2024} or 1-bit \cite{Zhang2025}), employing mixed-precision techniques \cite{Nguyen2021,Montgomerie2023}, pruning model parameters to reduce computational load and memory usage, or knowledge distillation to transfer knowledge from the quantized model into an even smaller and faster student model.

\subsection{Towards practical on-satellite deployment}
First, our vessel detection implementation would also need to be deployed alongside low-power and fast (\textless 1 minute) FPGA SAR processing (as already demonstrated by \cite{Kerr2020, Baungarten2023,Breit2021}), which converts raw SAR data (L0) to the rasterized scenes (L1) needed as input.
In addition, the detection threshold would need to be adjusted to reflect specific mission requirements. Here we adopt a relatively conservative threshold, which produces the best results in evaluation using the xView3-SAR competition metrics, and results in fewer false positives (FP: 10,780 vs. 12,369) but more false negatives (FN: 15,670 vs. 13,345) in comparison to the winning model. However, where minimization of missed detections is critical, optimizing the threshold for high recall (lower false negative rate) would be more suitable, accepting the increased computational cost of processing additional candidate detections.

We also note that limited on-satellite storage may place constraints on the SAR processing and analysis pipeline, particularly for the bathymetry data used here as an auxiliary input data channel. These data are not large (global dataset 7.5 GB, 4 GB compressed \cite{2025GEBCO}) in comparison to storage capacity of large satellites (Pléiades Neo has 1.75 TB), so could be stored on space-hardened storage, periodically updated from ground stations, and co-registered with observation areas using onboard logic. Alternatively, the bathymetry could be omitted at a minor cost to performance (reduction of classification and near-shore detection by 1.3\% on average; see ablation results in Table \ref{tab:model_ablations}.B), or could be replaced with alternative contextual information generated onboard from SAR data (e.g. land masks from thresholding using Otsu's method \cite{Ji2016}), at the cost of additional onboard compute.

\section{Summary}
In this study we have demonstrated that highly-performant ML SAR vessel detection is possible in power-restricted on-satellite systems, with the quasi real-time latency (\textless 1 min for a $\sim$40,000 km$^2$ SAR scene) required for real-world maritime intelligence and satellite autonomy applications. We have achieved this through systematic exploration and combination of previous CNN model adaptations to create a tailored model for this task, and through deployment and testing of our new model on the Kria KV260 platform, operating within a CubeSat-relevant sub-10W power budget. In the long term, development of on-board ML has potential to transform satellite constellations from passive data collection systems into active monitoring networks capable of immediate response to events of interest; ultimately enhancing global security, environmental protection, and disaster response.

\begin{acks}
This work was supported by the Turing's Defence and Security programme through a partnership with Dstl. This research used the Baskerville Tier 2 HPC; funded by EPSRC \& UKRI (EP/T022221/1, EP/W032244/1) and operated by Advanced Research Computing, U. Birmingham. Copernicus Sentinel-1 SAR data are provided courtesy of ESA.
\end{acks}

\bibliographystyle{ACM-Reference-Format}
\bibliography{references}

\end{document}